\documentclass{article}

\usepackage{arxiv}

\usepackage[utf8]{inputenc} 
\usepackage[T1]{fontenc}    
\usepackage{hyperref}       
\usepackage{url}            
\usepackage{booktabs}       
\usepackage{amsfonts}       
\usepackage{nicefrac}       
\usepackage{microtype}      
\usepackage{lipsum}		
\usepackage{graphicx}
\usepackage{natbib}
\usepackage{doi}
\usepackage{graphicx}
\usepackage{subcaption}

\title{Large language models surpass domain-specific architectures for cardiotocography}

\author{
    Sheng Wong \\
    Nuffield Department of Women's \& Reproductive Health\\
    University of Oxford\\
    Oxford, United Kingdom \\
    \And
    Ravi Shankar \\
    Nuffield Department of Women's \& Reproductive Health\\
    University of Oxford\\
    Oxford, United Kingdom \\
    \And
    Beth Albert \\
    Nuffield Department of Women's \& Reproductive Health\\
    University of Oxford\\
    Oxford, United Kingdom \\
    \And
    Gabriel Davis Jones\thanks{Corresponding author: gabriel.jones@ox.ac.uk} \\
    Nuffield Department of Women's \& Reproductive Health\\
    University of Oxford\\
    Oxford, United Kingdom
}

\renewcommand{\headeright}{}
\renewcommand{\undertitle}{}
\renewcommand{\shorttitle}{}

\hypersetup{
pdftitle={LLMs surpass domain-specific architectures for cardiotocography},
pdfsubject={q-bio.NC, q-bio.QM},
pdfauthor={Sheng Wong},
pdfkeywords={Electronic fetal monitoring, Cardiotocography, Obstetrics, Time-Series Classification, Large Language Models, Foundation Models},
}

\begin{document}
\maketitle

\begin{abstract}
Foundation models (FMs) and large language models (LLMs) have demonstrated promising generalization across diverse domains for time-series analysis, yet their potential for electronic fetal monitoring (EFM) and cardiotocography (CTG) analysis remains underexplored. Most existing CTG studies relied on domain-specific models and lack systematic comparisons with modern foundation or language models. limiting our understanding of whether these models can outperform specialized systems in fetal health assessment. In this study, we present the first comprehensive benchmark of state-of-the-art architectures for automated antepartum CTG classification. Over 2,500 20-minutes recordings were used to evaluate over 15 models spanning domain-specific, time-series, foundation, and language-model categories under a unified framework. Fine-tuned LLMs consistently outperformed both foundation and domain-specific models across data-availability scenarios, except when uterine-activity signals were absent, where domain-specific models showed greater robustness. These performance gains, however, required substantially higher computational resources. Our results highlight that while fine-tuned LLMs achieved state-of-the-art performance for CTG classification, practical deployment must balance performance with computational efficiency.

\end{abstract}


\section{Introduction}
Cardiotocography (CTG) monitoring is an integral component of antepartum care, providing clinicians with continuous assessment of fetal well-being through simultaneous recording of fetal heart rate (FHR) and uterine activity (UA) using a tocodynamometer (TOCO) \cite{cohen2012accuracy}. CTG recordings capture complex temporal patterns across multiple time scales, including heart rate variability, transient accelerations and decelerations, and their relationship to uterine contractions \cite{jones2022computerized}. Because these recordings often span 20 minutes to several hours, accurate interpretation requires careful analysis of temporal trends and signal characteristics unique to each fetus.

In current clinical practice, CTG assessment is performed visually by trained clinicians following established guidelines, an approach that is inherently subjective and time-consuming. Substantial inter and intra-observer variability has been documented, with studies showing poor agreement even among experienced practitioners when reviewing the same CTG traces \cite{gagnon1993comparison,rei2016interobserver,hernandez2023reliability,tolladay2025comparing}. This inconsistency contributes to delayed interventions in cases of fetal distress as well as unnecessary interventions such as avoidable cesarean deliveries. These limitation has underscore the critical need for robust automated analysis methods that can provide consistent, objective, and reproducible assessments of fetal well-being to support clinical decision-making.

Early automation of CTG assessment relied on features engineering, but recent studies have proposed domain-specific deep learning (DL) architectures for antepartum CTG to identify fetal well-being \cite{khan2025patchctg, wong2025cleanctg,chen2021intelligent,chiou2025development}. These models learn to automatically identify distinct characteristics and patterns from raw CTG signals with minimal preprocessing, achieving state-of-the-art (SOTA) results in their studies. Nevertheless, they have yet to leverage the representational learning capabilities and transfer learning potential offered by large-scale models. 

In parallel, large language models (LLMs) and foundation models (FMs) have demonstrated unprecedented performance across multiple domains, achieving capabilities that were previously thought impossible \cite{awais2025foundation, naveed2025comprehensive}. These models, trained on billions of parameters and large corpus of publicly available data, have demonstrated impressive capabilities in various tasks, including natural language understanding, computer vision, and time-series forecasting \cite{liang2024foundation,naveed2025comprehensive}. The primary advantage of these models lies in their ability to learn dense, generalizable representations of real-world phenomena, enabling effective adaptation to multiple downstream tasks through human-interpretable instructions. In healthcare, they have demonstrated proficiency in diagnostic reasoning, medical question answering, and medical imaging analysis \cite{garcia2024medical,shool2025systematic,panagoulias2024evaluating,bachmann2024exploring, penny2025reducing}. More recently, their applications and extensions to complex time-series data, including physiological signals such as EEG and EMG, have demonstrated robust performance and strong pattern recognition capabilities \cite{liang2024foundation,jin2023time,feofanov2025mantis,kim2024health}. 

Despite this advancement, investigation into model architectures such as LLMs and FMs for CTG analysis remains limited, often constrained by small datasets and lack comprehensive evaluation \cite{sun2025ctg,psilopatis2025comparative,GUMILAR20251140}. Furthermore, CTG-specific models have yet to be evaluated alongside recent advances in general-purpose time-series and FMs within a unified experimental framework. The absence of systematic comparisons and comprehensive exploration of SOTA model architectures leaves the potential of modern AI approaches largely unrealized, depriving clinicians and researchers of clear guidance on the most effective methods for automated fetal monitoring. 

To address these gaps, we present the first benchmark of SOTA architectures for antepartum CTG classification. We evaluated over 15 model architectures spanning domain-specific models, general-purpose time-series models, time-series FMs, and LLMs, using 2,500 real-world 20-minute CTG recordings. All models are trained or fine-tuned, and evaluated under a unified evaluation framework to ensure fair comparison. The analysis assesses performance across varying data-availability scenarios, robustness to absent UA, and computational efficiency. This benchmark provides an objective performance reference for CTG classification and offers evidence-based insights into the strengths and limitations among modern AI architectures for fetal monitoring. 

\section{Results}
We evaluated 14 model architectures with 18 variants across 4 categories: domain-specific DL models, general-purpose time-series models, time-series FMs, and LLMs. The evaluation encompassed PatchCTG, Conv-PatchCTG, 1D SE-ResNet, and NeuroFetalNet as CTG-specific baselines, 3 general-purpose time-series transformers (Informer, Non-stationary Transformer, TimesNet), two time-series FMs (Mantis and Moment), and multiple LLMs including Llama 1B–8B variants, Time-LLM adaptations, and GPT-5-mini with distinct prompting strategies. Model responses were evaluated using the same training, validation, and testing splits to ensure comparability. All models were trained or fine-tuned to classify fetal outcomes as normal pregnancy outcomes (NPO) or abnormal pregnancy outcomes (APO) based on real-world 20-minute CTG recordings from the Oxford Maternity (OxMat) database under multiple experimental scenarios. Notably, this dataset has not been publicly released, enabling us to evaluate the true generalization capabilities of all fine-tuned FMs and LLMs without concerns about potential data leakage during their pre-training phases. Detailed performance metrics, including Area under the Curve (AUC), accuracy, sensitivity and specificity for all models across experimental scenarios are provided in Supplementary S1. The models, dataset as well as data preprocessing are explained in the Methods section below.

\subsection{Overall Performance for CTG classification}
\begin{figure}[!htbp]
    \centering
    \includegraphics[width=1\textwidth]{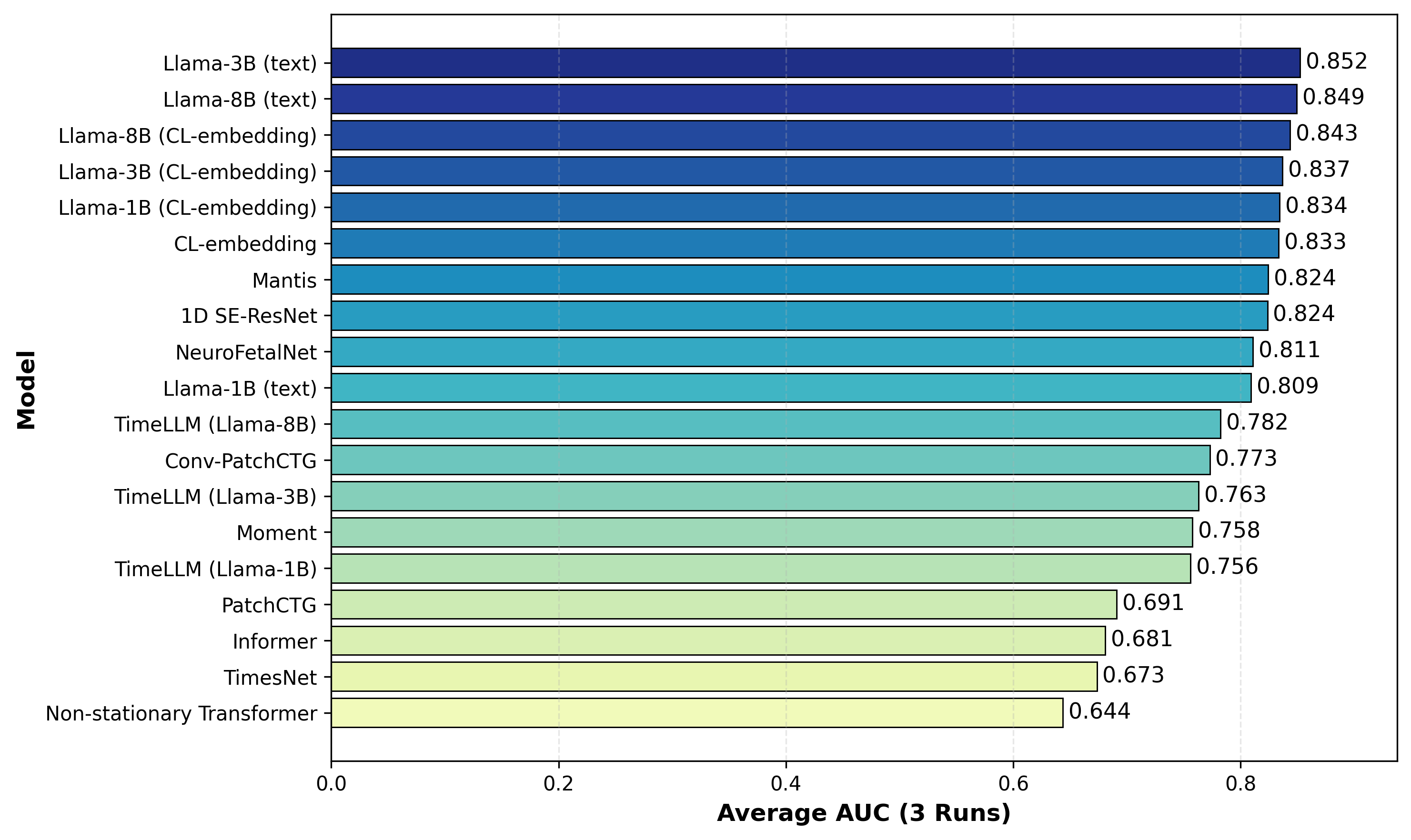}
    \caption{Average AUC performance for all models}
    \label{fig:validate}
\end{figure}

The performance of all models is presented in Figure \ref{fig:validate}. fine-tuned Llama-3B (text) performed the best, followed by Llama-8B (text) and Llama-3B (CL-embedding) at 0.852, 0.849, and 0.837 respectively. Interestingly, models using CL-embedding for pretraining: Llama-3B (CL-embedding) and Llama-1B (CL-embedding) performed similarly, suggesting that contrastive learning alone can achieve comparable performance. The best domain-specific architecture, 1D SE-ResNet, achieved identical performance to the best time-series FM, Mantis, at 0.833. We also observed that the worst-performing models were predominantly general time-series models, with Non-stationary Transformer scoring the lowest with an AUC of 0.644. Additionally, PatchCTG, which is a domain-specific architecture, performed similarly to Informer and did not come close to other domain-specific models, with an AUC of 0.691. However, the convolutional variant, Conv-PatchCTG, showed remarkably improved performance of 0.773, demonstrating that incorporating convolutional layers substantially enhances model performance.

\begin{figure}[!htbp]
    \centering
    \includegraphics[width=1\textwidth]{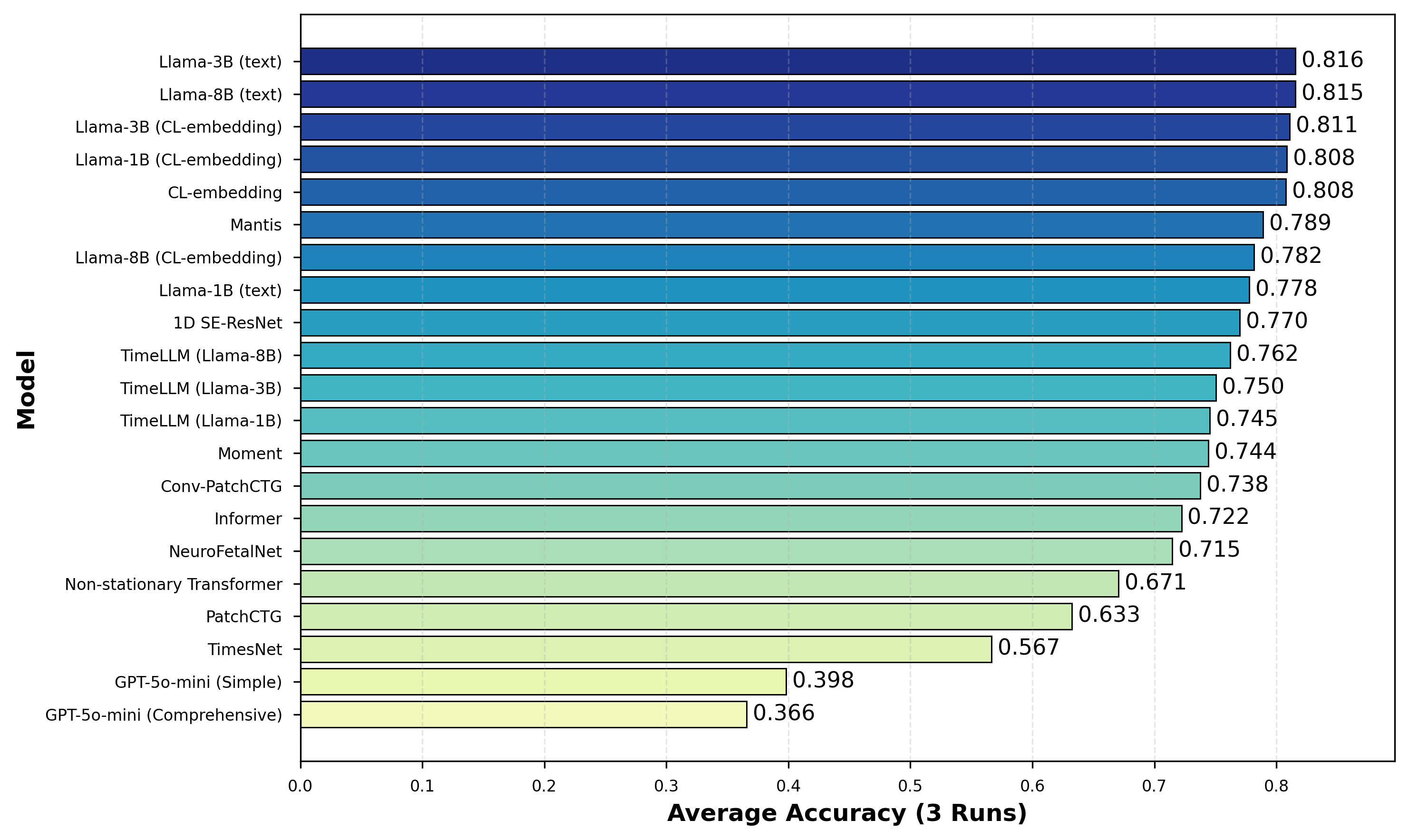}
    \caption{Average accuracy for all models}
    \label{fig:validate acc}
\end{figure}

Furthermore, we evaluated all models using accuracy as a secondary metric to enable comparison with the untrained LLM GPT-5-mini, which does not output probabilistic predictions. As shown in Figure \ref{fig:validate acc}, both simplified and detailed one-shot prompting strategies yielded no discriminative ability for CTG classification, with performance substantially worse than all other models and a strong bias toward positive predictions. Additionally, more detailed and comprehensive prompting strategies did not improve accuracy. Given this poor baseline performance, we excluded GPT-5-mini from further investigation and ablation studies, focusing our analysis on models that demonstrated meaningful discriminative capability.

\subsection{Data-constrained Scenario}
\begin{figure}[!htbp]
    \centering
    \includegraphics[width=1\textwidth]{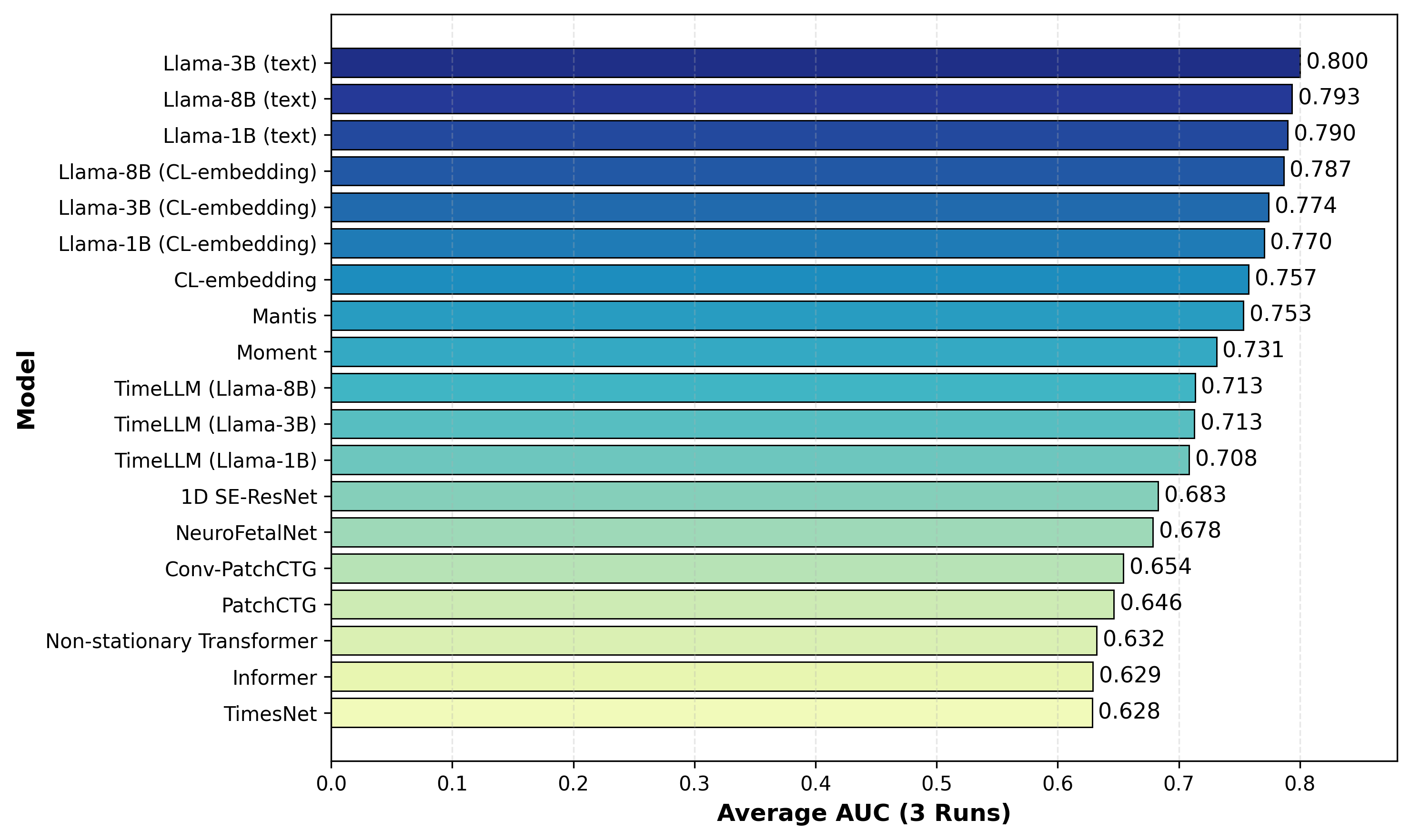}
    \caption{Average AUC performance across all models trained with limited training data}
    \label{fig:limited data}
\end{figure}

When evaluating model performance under data-constrained conditions, as detailed in Figure \ref{fig:limited data}, all architectures experienced performance degradation compared to full dataset training. The reduction in training data from the complete dataset of over 23,000 to 4,000 recordings resulted in AUC decreases of approximately 4\% - 6\% across majority of the models, reflecting the expected impact of reduced training data on model generalization. However, there were notable exceptions: domain-specific models such as Conv-PatchCTG, NeuroFetalNet, and 1D SE-ResNet suffered substantial decrease in AUC within the range of 12\% to 14\%, attributed to training from scratch compared to the fine-tuning process employed in LLM-based models.

The best-performing model architectures remain LLM-based model, with fine-tuned Llama-3B (text) achieving an AUC of 0.800, followed by Llama-8B (text) at 0.793 and Llama-1B (text) at 0.790, maintaining a performance advantage over other architectures and models. Additionally, Moment, a foundation model, maintained relatively stable performance despite reduced training data, decreasing by only 2\% in AUC. This robustness suggests that architectures pre-trained on large out-of-domain datasets remain relatively stable, with overall performance superior to domain-specific models and general time-series models under these conditions. There is an exception, however, CL-embedding model, achieved an AUC of 0.757, surpassing TimeLLM and fine-tuned time-series FMs.

\subsection{Uterine Activity (UA) Signal Ablation}
\begin{figure}[!htbp]
    \centering
    \includegraphics[width=1\textwidth]{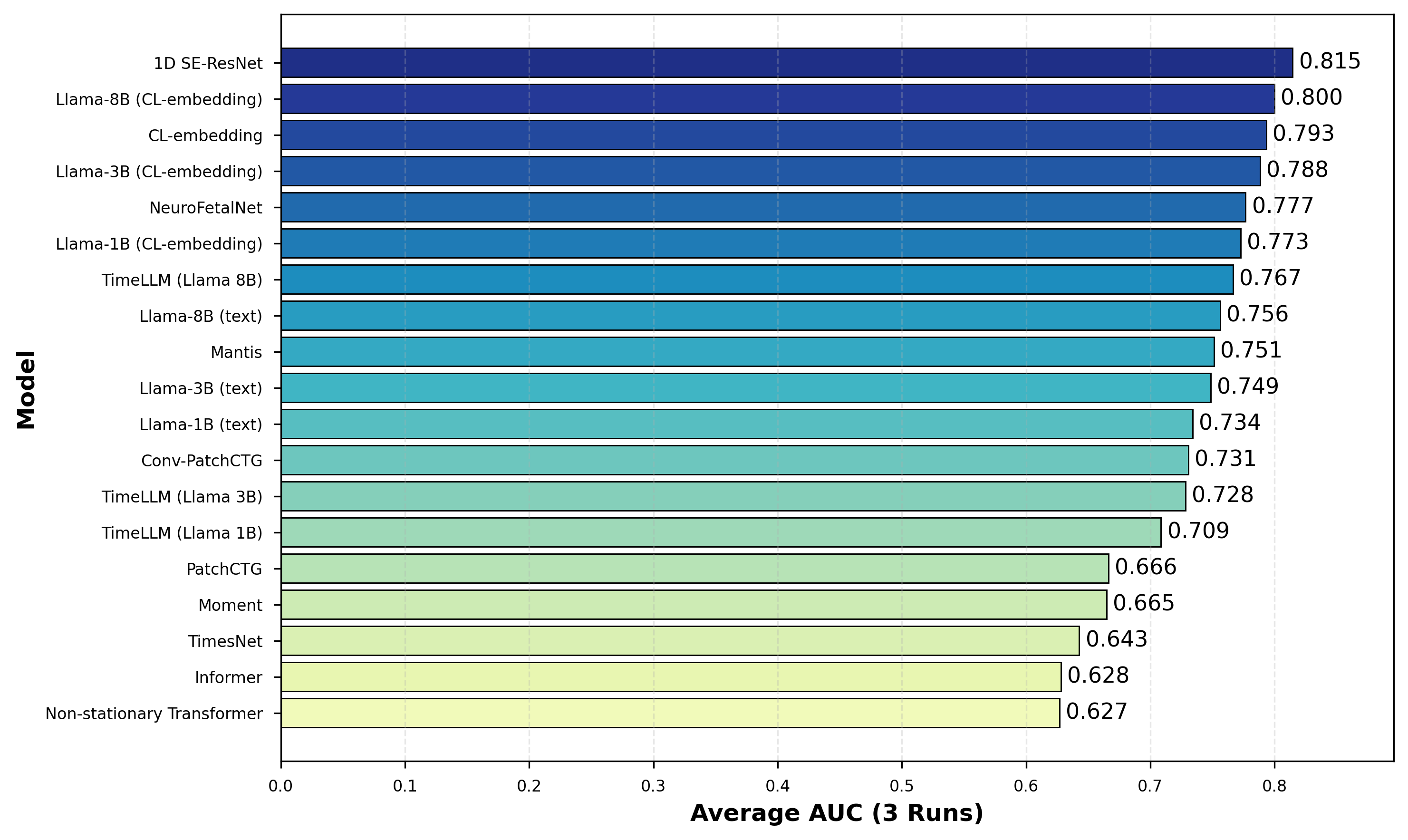}
    \caption{Average AUC performance across all models without UA}
    \label{fig:TOCO}
\end{figure}

The impact of missing UA signals for CTG on model performance is presented in Figure \ref{fig:TOCO}. When UA signals were absent during inference for models originally trained with complete FHR and UA, all architectures experienced performance degradation, with AUC reductions ranging from 3\% to 10\%. This degradation demonstrates that uterine contraction patterns provide a minor to moderate contribution to accurate CTG classification across all modeling paradigms. 1D SE-ResNet achieved the best performance with an AUC of 0.815, followed by Llama-8B (CL-embedding) at 0.800 and CL-embedding at 0.797. Notably, text-based Llama models experienced the most substantial performance degradation, with Llama-8B (text) declining from 0.852 with UA signals to 0.749 without them, a drop of around 10.3\%.

Models incorporating convolutional layers, including CL-embedding variants, their LLM-backbone counterparts, and domain-specific architectures such as 1D SE-ResNet and NeuroFetalNet demonstrated greater robustness to missing UA data. These CNN-based models consistently outperformed text-based Llama models in the absence of UA signals, exhibiting smaller performance degradation and maintaining higher overall performance. This suggests that convolutional architectures provide inherent advantages for handling incomplete multivariate time-series data, likely due to their channel-based processing.

\subsection{Temporal Dependency Analysis}
\begin{figure}[!htbp]
    \centering
    \includegraphics[width=1\textwidth]{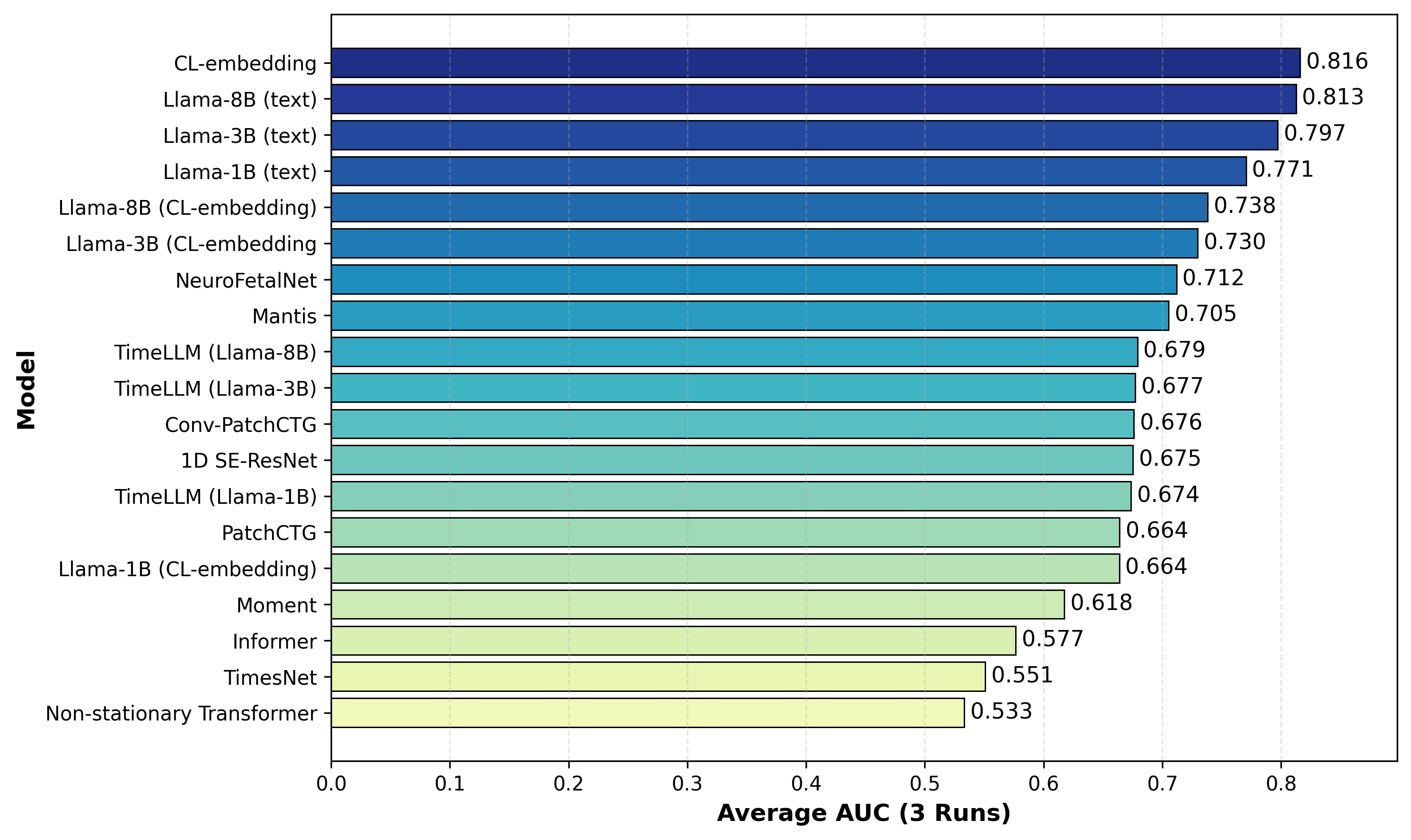}
    \caption{Average AUC performance across all models with temporal shuffling}
    \label{fig:temporal_dependency}
\end{figure}

The performance of models under temporal shuffling is presented in Figure \ref{fig:temporal_dependency}. Following temporal shuffling, standalone CL-embedding achieved the highest AUC at 0.816, followed by Llama-8B (text) at 0.813 and Llama-3B (text) at 0.797. All models experienced performance degradation compared to the original non-shuffled data.
Text-based Llama models showed moderate degradation from temporal shuffling: Llama-8B (text) declined by 4.2\% (from 0.849 to 0.813), Llama-3B (text) by 6.5\% (from 0.852 to 0.797), and Llama-1B (text) by 4.3\% (from 0.809 to 0.774). Standalone CL-embedding exhibited the smallest decline at 2.0\% (from 0.833 to 0.816).

In contrast, models integrating CL-embedding with LLM backbones experienced substantially larger degradation. Llama-1B (CL-embedding) showed the greatest decline from 0.834 to 0.664). TimeLLM variants also showed significant degradation, declining 11.4\% for Llama-8B backbone, 11.7\% for Llama-3B backbone, and 10.9\% for Llama-1B backbone. Domain-specific architectures demonstrated comparable vulnerability to temporal disruption. 1D SE-ResNet declined by 18.1\% (from 0.824 to 0.675), NeuroFetalNet by 12.2\% (from 0.811 to 0.712), and Conv-PatchCTG by 12.6\% (from 0.773 to 0.676). FMs showed similar patterns, with Moment declining 18.5\% (from 0.758 to 0.618) and Mantis 14.4\% (from 0.824 to 0.705). Traditional time-series models also experienced major reduction in performance ranging from 15.3\% to 18.1\% across Informer, Non-stationary Transformer, and TimesNet architectures.

\subsection{Discussion}
In this study, we conducted a comprehensive evaluation of SOTA approaches for antenatal CTG analysis, comparing DL models, time-series FMs, and multiple variants of LLMs across a dataset of 2,500 CTG recordings. Our findings demonstrate that fine-tuned LLMs with text-based input consistently outperform both domain-specific architectures and foundation models, achieving an AUC of 0.852 with Llama-3B, representing minor to moderate improvement over previously reported performance metrics in CTG classification.

The superior performance of fine-tuned LLMs persisted under scenarios with limited data. All LLM variants and time-series FMs outperformed domain-specific architectures in this scenario, suggesting that models leveraging pre-trained representations from large-scale datasets provide advantages when domain-specific training data are scarce. However, ablation studies examining model performance without UA revealed contrasting patterns. LLMs with text as inputs such as Llama-8B demonstrated substantial performance degradation compared to embedding-based LLM variants that utilize learned temporal representations as input to the LLM backbone. Conversely, domain-specific convolutional architectures, including 1D SE-ResNet and NeuroFetalNet, maintained relatively strong performance and outperformed several LLM variants and time-series foundation models under these conditions.

Analysis of model behavior under temporal shuffling revealed that most architectures experienced performance degradation of approximately 10\% or greater in AUC. Notably, text-based Llama models exhibited substantially smaller declines of 3-6\%, suggesting reduced dependency on precise temporal ordering. This enhanced robustness may reflect the models capacity to learn from distributional characteristics of CTG signals, such as heart rate range and variability metrics, rather than relying primarily on sequential temporal patterns.

It is important to note that the observed performance decline under temporal shuffling may not exclusively reflect temporal dependency learning. The shuffling process potentially disrupts both temporal structure and non-temporal statistical patterns that models utilize for classification, making it difficult to isolate the specific contribution of sequential information. Therefore, while our results strongly suggest that models learn temporal features, we cannot definitively attribute all performance degradation solely to the loss of temporal ordering, as disruption of aggregate statistical properties may also contribute to the observed declines.

We also evaluated GPT-5-mini using one-shot prompting with two distinct instruction strategies, which yielded particularly revealing results. Despite being a SOTA LLM, GPT-5-mini demonstrated no discriminative ability for CTG classification, achieving performance substantially worse than all the other models. This finding contrasts sharply with studies reporting successful CTG interpretation using LLMs. However, it should be noted that our evaluation dataset of 2,500 recordings substantially exceeds the sample sizes used in those studies, which typically utilized fewer than 100 recordings \cite{sun2025ctg,psilopatis2025comparative,GUMILAR20251140}. The observed discrepancy suggests that previously reported positive results may not generalize to larger, more diverse clinical datasets with greater signal variability and noise characteristics typical of real-world deployment scenarios. Furthermore, methodological differences across studies complicate direct comparisons, as previous evaluations have employed varying approaches including image-based CTG representations and multi-agent prompting frameworks, whereas our evaluation used direct numerical time-series input. 

\begin{figure}[!htbp]
    \centering
    \begin{subfigure}[b]{1\textwidth}
        \centering
        \includegraphics[width=\textwidth]{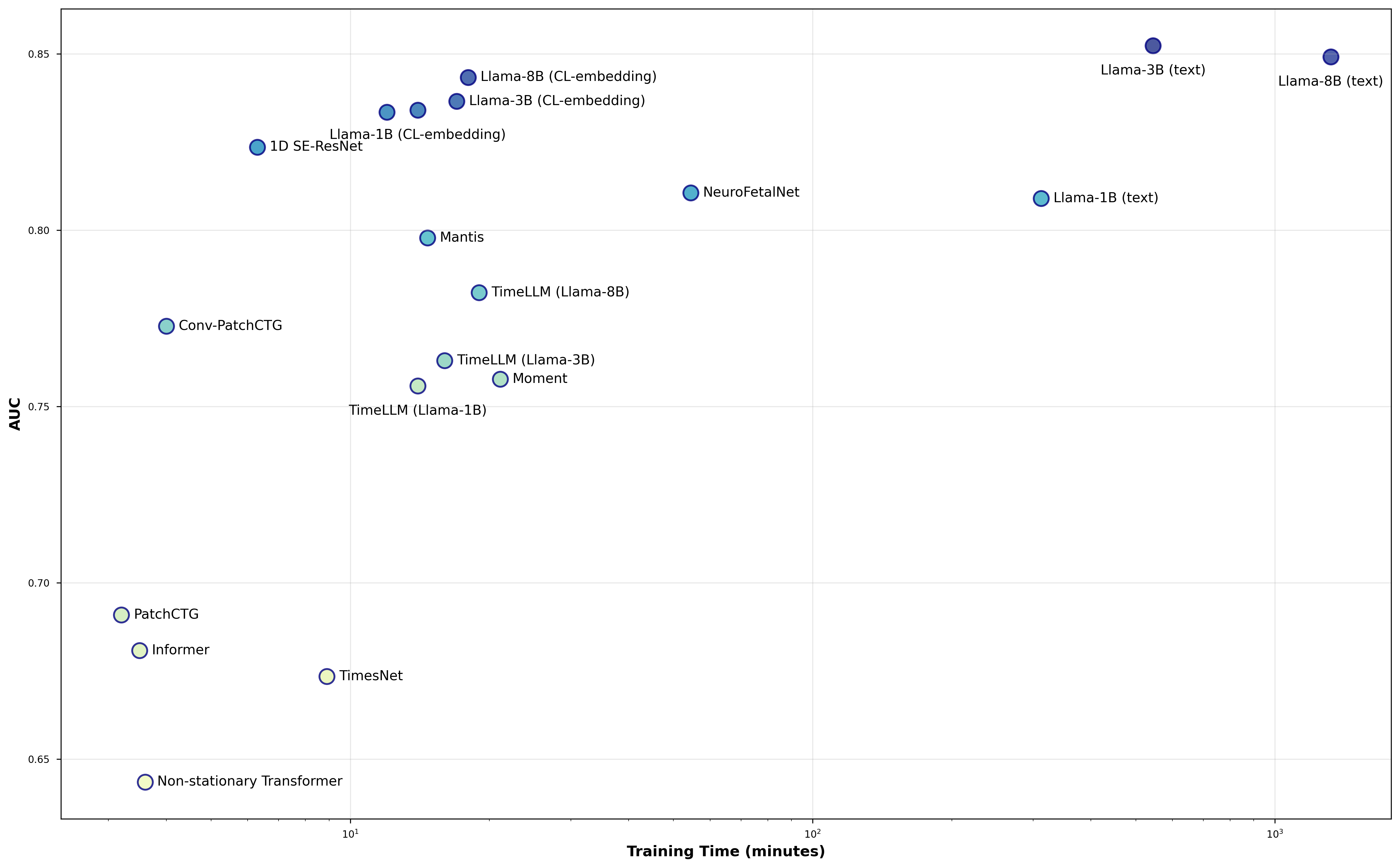}
        \caption{Training time}
        \label{fig:training speed}
    \end{subfigure}
    
    \vspace{0.5cm}
    
    \begin{subfigure}[b]{1\textwidth}
        \centering
        \includegraphics[width=\textwidth]{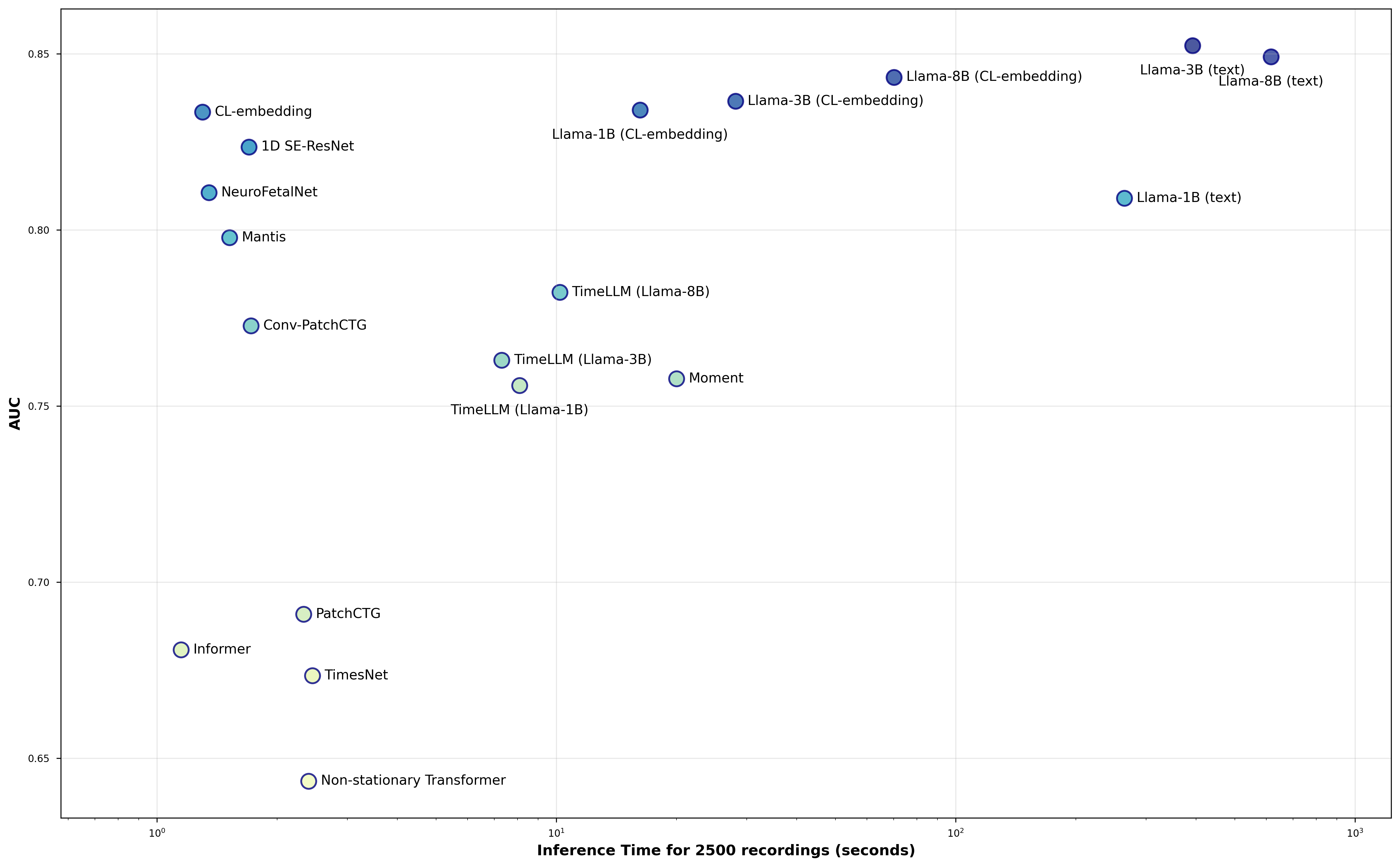}
        \caption{Inference time}
        \label{fig:inference speed}
    \end{subfigure}
    \caption{(a) Shows AUC versus model parameters for complete bivariate CTG input. (b) Shows performance degradation when uterine activity signals are absent.}
    \label{fig:2}
\end{figure}

Additionally, the computational requirements for processing 20 minutes CTG recordings presents significant computation challenges. Our approach of using stride-based sampling to reduce context length still required fine-tuning on up to 3 consumer-grade GPUs (RTX 5090) for text-based Llama models. The modest improvement of 2\%-3\% in AUC achieved by LLM variants came at a computational cost several-fold greater than domain-specific architectures. Fine-tuning for Llama-8B (text) exceeded 22 hours, while CL-embedding and 1D SE-ResNet both required less than 15 minutes, despite Llama-8B achieving only a 2\%-3\% improvement in AUC over these architectures. This substantial disparity in computational efficiency is illustrated in Figure \ref{fig:training speed} and Figure \ref{fig:inference speed}, which compare training and inference speeds across all evaluated architectures.

Our preliminary experiments with LLMs without fine-tuning revealed severe limitations, including frequent hallucinations and inconsistent outputs that rendered them unsuitable for evaluation or analysis due to the longer context length. This finding reinforces the necessity of supervised fine-tuning such as LORA for medical applications, even when using SOTA pre-trained models. For clinical users with limited computational resources, domain-specific architectures may therefore represent a more pragmatic choice, offering competitive performance with substantially lower infrastructure requirements and faster implementation cycles. 

While we evaluated over 15 model architectures, our study does not encompass the full landscape of publicly available models. Other LLMs such as Qwen and Phi, as well as additional foundation models, were not included in our analysis due to computational constraints and time limitations. We selected best representative models from each architectural landscape to provide meaningful comparisons while maintaining feasibility within our computational budget. Additionally, our hyperparameter optimization was intentionally limited to maintain computational feasibility across multiple architectures. We primarily relied on published hyperparameters (when available) from original model implementations, which may not represent optimal configurations for our CTG data. More extensive architecture-specific tuning, particularly for domain-specific models could potentially narrow the observed performance gaps. 

Overall, fine-tuned LLMs achieved higher discriminative performance than time-series FMs and CTG-specific architectures, with the best AUC reaching 0.852. These results position fine-tuned LLMs as a credible option for automated CTG analysis. However, clinical use will require prospective, multicentre validation; assessment of calibration, decision thresholds, and false positive burden; monitoring for distribution shift; and evaluation of equity, safety, and cost-effectiveness. Reporting should include rigorous error analyses and decision-analytic impact, not accuracy alone. Future work should test parameter-efficient adaptation methods for more advanced reasoning models, explore multimodal integration with clinical covariates, and compare human–AI teaming against clinician-only review in pragmatic trials. Until such evidence is available, LLMs should be considered decision-support tools rather than replacements for expert interpretation.

\section{Methodology}
\subsection{Data Preparation}
The CTG recordings were obtained from the Oxford Maternity (OxMat) database, a longitudinal repository of antepartum CTGs collected at the John Radcliffe Hospital, Oxford, UK, between 1991 and 2024 \cite{khan2024oxmat}. The CTG recordings contain FHR and UA signals of varying lengths. The dataset includes over 250 clinical variables encompassing maternal and neonatal outcomes such as Apgar scores, cord blood gas values, birthweights, and delivery types. The database has been extensively used in various CTG research tasks.  

Based on these clinical variables, CTG signals were categorized into two cohorts: Adverse Pregnancy Outcome (APO) and Normal Pregnancy Outcome (NPO). The adverse pregnancy outcome cohort included recordings from fetuses with one or more of the following outcomes: antepartum or intrapartum stillbirth, neonatal death within 24 hours of birth, or fetal asphyxia. Selection criteria required recordings to have less than 50\% missing data and a minimum duration of 10 minutes. To reflect real-world clinical settings, recordings were preserved in their original lengths and segmented into 20-minute windows without additional preprocessing or cleaning. All CTG signals were downsampled from 4 Hz to 1 Hz to standardize computational requirements. Due to the long context window required for 20 minutes of text-based input, we further downsampled the data to 0.5Hz to reduce computational overhead and mitigate potential issues with long sequence processing in LLM-based models. This approach enabled model fine-tuning on 3 32GB RTX 5090 GPUs. 

The segmented data were then split into training and evaluation sets. The training set contained 23,127 20-minutes CTG segments: 16,526 NPOs and 6,601 APOs. The evaluation set contained 2,500 segments: 1,800 NPOs and 700 APOs, yielding an approximate training-to-evaluation ratio of 9.25:1. The complete cohort development has been previously described \cite{davis2025performance}.

\subsection{Model Selection \& Setup}
\begin{table}[!ht]
    \centering
    \resizebox{\textwidth}{!}{%
    \begin{tabular}{lccccc}
    \toprule
       \textbf{Model} & \textbf{Trainable Parameters} & \textbf{Model Category} & \textbf{Training Method} & \textbf{Input Format} & \textbf{Sampling Rate} \\
    \midrule
        \multicolumn{6}{l}{\textit{Domain-Specific Models}} \\
        \quad PatchCTG & 7.3M & CTG-specific & Trained from scratch & Structured time-series & 1 Hz \\
        \quad Conv-PatchCTG & 7.3M & CTG-specific & Trained from scratch & Structured time-series & 1 Hz \\
        \quad 1D SE-ResNet & 45M & CTG-specific & Trained from scratch & Structured time-series & 1 Hz \\
        \quad NeuroFetalNet & 2.4M & CTG-specific & Trained from scratch & Structured time-series & 1 Hz \\
    \midrule
        \multicolumn{6}{l}{\textit{General-Purpose Time-Series Models}} \\
        \quad Informer & 1.1M & Time-series transformer & Trained from scratch & Structured time-series & 1 Hz \\
        \quad Non-stationary Transformer & 0.4M & Time-series transformer & Trained from scratch & Structured time-series & 1 Hz \\
        \quad TimesNet & 8.7M & Time-series CNN & Trained from scratch & Structured time-series & 1 Hz \\
    \midrule
        \multicolumn{6}{l}{\textit{Time-Series Foundation Models}} \\
        \quad Moment & 0.3B & Pre-trained time-series FM & Full fine-tuning & Structured time-series & 1 Hz \\
        \quad Mantis & 8M & Pre-trained time-series FM & Full fine-tuning & Structured time-series & 1 Hz \\
    \midrule
        \multicolumn{6}{l}{\textit{Large Language Models}} \\
        \quad Llama 3 (1B) & 9M & LLM & QLoRA fine-tuning & Text (full-length CTG)) & 0.5 Hz \\
        \quad Llama 3 (3B) & 11M & LLM & QLoRA fine-tuning & Text (full-length CTG)) & 0.5 Hz \\
        \quad Llama 3 (8B) & 13M & LLM & QLoRA fine-tuning & Text (full-length CTG)) & 0.5 Hz \\
        \quad GPT-5-mini (Simple) & Unknown & Proprietary LLM & One-shot prompting & Text (full-length CTG) & 1 Hz \\
        \quad GPT-5-mini (Detailed) & Unknown & Proprietary LLM & One-shot prompting & Text (full-length CTG) & 1 Hz \\
    \midrule
        \multicolumn{6}{l}{\textit{Time-LLM Hybrid}} \\
        \quad Time-LLM (Llama 1B) & 0.1B & LLM hybrid & Frozen LLM + alignment training & Structured + text & 0.5 Hz \\
        \quad Time-LLM (Llama 3B) & 0.1B & LLM hybrid & Frozen LLM + alignment training & Structured + text & 0.5 Hz \\
        \quad Time-LLM (Llama 8B) & 0.1B & LLM hybrid & Frozen LLM + alignment training & Structured + text & 0.5 Hz \\
    \midrule
        \multicolumn{6}{l}{\textit{Contrastive Learning (CL) Models}} \\
        \quad CNN Encoder (CL) & 1.5M & ResNet-style CNN & Contrastive pre-training + classification & Structured time-series & 0.5 Hz \\
        \quad Llama 1B (CL-embedding) & 22M & LLM hybrid  & CL pre-training + QLoRA fine-tuning & Structured embeddings & 0.5 Hz \\
        \quad Llama 3B (CL-embedding) & 48M & LLM hybrid & CL pre-training + QLoRA fine-tuning & Structured embeddings & 0.5 Hz \\
        \quad Llama 8B (CL-embedding) & 58M & LLM hybrid & CL pre-training + QLoRA fine-tuning & Structured embeddings & 0.5 Hz \\
    \bottomrule
    \end{tabular}%
    }
    \vspace{0.2cm}
    \caption{Model specifications and training methodologies for experimental comparison.}
    \label{tab:model_specifications}
\end{table}

We evaluated 14 model architectures with 18 model variants across 4 categories: domain-specific DL models, general-purpose time-series models, time-series FMs, and LLMs. These variants included different model sizes, prompting strategies GPT-5-mini with simple and detailed instructions, and input modalities (For example: structured embeddings, CL-embedding and text). Table \ref{tab:model_specifications} summarizes model architectures and parameter counts, which range from thousands to 8 billion parameters for publicly available models. Hyperparameters for each architecture were maintained as specified in their original publications without additional tuning, except otherwise stated below.

For traditional DL approaches, we selected \textbf{PatchCTG} \cite{khan2025patchctg}, and it's variant \textbf{Conv-PatchCTG} \cite{khan2025patchctg}, \textbf{1D SE-ResNet} \cite{park2025automated} and \textbf{NeuroFetalNet} \cite{sun2024neurofetalnet} as domain-specific baselines. These models were recently published in 2025 for CTG analysis, representing the current SOTA in domain-specific CTG analysis approaches. PatchCTG was chosen because it was trained and evaluated on a CTG dataset from the same database as our study, enabling direct performance comparison. 1D SE-ResNet was selected due to its validation on multiple external datasets, demonstrating strong generalizability across different hospital settings. NeuroFetalNet was included because it achieved SOTA performance when compared against other existing models in its evaluation. Most importantly, these architectures are publicly available and can be adapted with minimal modifications to our evaluation setup, ensuring easier reproducibility and a fairer evaluation. \textbf{Informer} \cite{zhou2021informer}, \textbf{Non-stationary Transformer} \cite{liu2022non} and \textbf{TimesNet} \cite{wu2022timesnet} as general-purpose time-series classification models. These architectures represent current lightweight approaches in CTG analysis with DL. Although these general-purpose models have demonstrated strong performance in other classification tasks, they have not yet been evaluated for CTG analysis.

Two recent time-series foundation models, \textbf{Mantis} \cite{feofanov2025mantis} and \textbf{Moment} \cite{goswami2024moment}, were selected based on their demonstrated performance in healthcare classification tasks. Both models were pre-trained on large-scale time-series datasets and have shown superior performance compared to LLM-based time-series methods. Mantis was proposed as a lightweight time-series foundation model specifically designed for classification tasks, demonstrating that efficient architectures can outperform heavier LLM-based approaches. Moment represents a family of foundation models designed for multiple time-series tasks including classification, and has similarly been shown to exceed the performance of LLM-based time-series methods while maintaining computational efficiency.

For LLMs, we evaluated 3 variants of \textbf{Llama 3 Models} \cite{touvron2023llama}: the 1B, 3B and 8B models. These relatively tiny models have become a standard benchmark and have been used for extensive validation and evaluation across various downstream applications, making them an ideal candidate for this study. These publicly available language models were adapted for CTG classification by removing the original language modeling head and replacing it with a binary classification head. These models receive time-series data as text inputs. 

Additionally, we adapted \textbf{Time-LLM} \cite{jin2023time}, a framework originally designed for reprogramming language models for time-series forecasting, by replacing the forecasting head with a binary classification head. The framework 'reprograms' time-series data into tokenized form and aligns these tokens with LLM embedding layers through a multi-head cross-attention mechanism, combined with a short descriptive prompt, while keeping the LLM backbone frozen. This parameter-efficient fine-tuning approach has been widely adopted in recent time-series analysis studies. We selected Llama 1B, 3B and 8B as the LLM backbones.

We also employed a CNN encoder consisting of a ResNet-style architecture with an initial convolutional layer (kernel=7, stride=2) followed by two residual blocks (kernel=3), with channel dimensions of 2, 64, 128, batch normalization, ReLU activations with residual skip connections, and global average pooling \cite{koonce2021resnet}. This encoder was pre-trained using supervised contrastive learning (CL), a technique that learns representations by pulling same-class samples together while pushing different class samples apart in embedding space \cite{khosla2020supervised}. The features extracted from this pre-trained encoder were then projected to the embedding layers of Llama 1B, 3B and 8B models for subsequent fine-tuning. We refer to these variants as \textbf{Llama Models (CL-embedding)}. This approach offers an alternative to Time-LLM by directly injecting learned temporal representations into the LLM embedding space, rather than relying on text-based tokenization and cross-attention alignment. Additionally, we included the pre-trained \textbf{CNN encoder (CL embedding)} with a classification head as a standalone baseline for ablation study, allowing us to isolate the contribution of contrastive learning independent of LLM integration and assess whether the LLM components provide additional performance gains beyond the learned temporal representations. 

We additionally evaluated \textbf{GPT-5-mini} using two distinct prompting strategies to assess the performance of larger-scale LLMs without fine-tuning. The prompts were authored by a medical doctor with expertise in CTG interpretation. The first strategy (GPT-5-mini (Detailed Instruction)) employed comprehensive prompting that incorporated detailed clinical guidelines and step-by-step CTG analysis instructions, while the second (GPT-5-mini (Simple Instruction)) utilised a simplified prompt containing only essential summarised classification criteria (see supplementary S2). Unlike other models in our evaluation, GPT-5-mini processed full-length CTG recordings without context reduction, as it operates through an online API with extended context capabilities. 

\subsection{Training \& Evaluation}
All models were either trained or fine-tuned for the CTG classification task, with the exception of GPT-5-mini, which was evaluated using one-shot prompting with authored instructions. Traditional DL models were trained from scratch on our training set, with 5\% reserved for validation. Training continued for up to 100 epochs with early stopping triggered when validation AUC failed to improve for five consecutive epochs. The learning rate was set to 0.001 with a batch size of 64, and the Adam optimizer was used during training.

Time-series FMs, Time-LLMs variants, and CL-embedding LLMs followed a similar training protocol with the same validation split and stopping criteria. The learning rate was set to 0.0001 with a batch size of 64, and the Adam optimizer was used during the fine-tuning/training process.

For LLMs with text as tokenized input (structured values converted to strings), we employed LoRA-based fine-tuning with a rank of 32 and scaling factor of 64, specifically adapting the query, key, value, and output projection layers \cite{hu2022lora}. The learning rate was set to 0.0001 with a batch size of 64, and the Adam optimizer was used during the fine-tuning/training process. To reduce computation overhead, these models were fine-tuned for 3 epochs using the same 5\% validation split for monitoring performance. Unlike traditional DL models where validation occurred after each epoch, validation for LoRA fine-tuning was performed every 100 training samples due to the faster convergence. All model utilsing The Llama models were quantized to 4-bit to reduce computational burden and memory requirements during training and fine-tuning. 

All models were trained to predict binary classification of CTG recordings as either APO or NPO. Model performance was evaluated using the Area Under the Receiver (AUC) as the primary metric, chosen for its ability to assess discriminative capability across different classification thresholds. Secondary metrics including sensitivity, specificity, and accuracy were computed to provide a comprehensive performance overview, with the results reported in Supplementary Material S1. Additionally, to ensure robustness and reliability of results, all models were trained and evaluated over 3 runs, and the reported metrics represent the average performance across these runs.

\subsection{Ablation Study}
To comprehensively evaluate the robustness of our models, we conducted three ablation studies examining different aspects of model performance under constrained conditions. 
\begin{enumerate}
    \item \textbf{Data-constrained Scenario:} We evaluated model performance on a reduced dataset of 4000 of 20-minutes CTG segments to simulate clinical scenarios where data availability may be limited. This is particularly relevant for smaller clinical facilities or resource-constrained settings where accumulating large datasets for model training may be impossible. By examining how different model architectures perform with reduced training size, we are able to assess which approaches maintain acceptable performance when data scarcity is a constraint.
    \item \textbf{Uterine Activity Signal Ablation:}  In clinical practice, UA signals frequently suffer from quality issues, including incomplete recordings or complete absence. To evaluate model robustness to missing UA information, we conducted an ablation test using only FHR signals for classification. This ablation study involved zeroing out all UA signals in the test data while using the originally trained models, thereby simulating scenarios where UA data becomes unavailable during inference. This approach allows us to assess whether models trained on complete CTG data can maintain diagnostic accuracy when UA signals are absent, reflecting real-world clinical situations where UA may fail or produce unreliable measurements after model deployment. 
    \item \textbf{Temporal Dependency Analysis:} To investigate the importance of temporal ordering in CTG analysis for evaluated models, we systematically disrupted the sequential structure of the recordings while preserving local patterns. Specifically, we segmented each CTG recording into 10-seconds chunks and randomly shuffled their order before model inference. We also applied random masking to segments of the CTGs to further test the models' dependence on continuous temporal information. This dual perturbation approach preserves short-term temporal features within each minute while eliminating long-range temporal dependencies across the recording. By comparing model performance on temporally shuffled and masked, versus intact sequences, we can quantify whether the architecture relies on long-range temporal patterns versus local feature detection. 
\end{enumerate}

\section{Funding}
This research was supported by the UKRI Medical Research Council (MR/X029689/1).

\section{Competing Interests}
All authors declare no financial or non-financial competing interests. 

\section{Data availability}
The dataset analysed during the current study are not publicly available due to privacy and ethical considerations, but are available from the corresponding author on reasonable request.
Further details about the dataset are available on the OXMAT website: https://www.oxdhl.com/resources.

\section{Code availability}
The code for NeruoFetalNet is available at https://github.com/BlackThompson/NeuroFetalNet. The code for TimeLLM is available at https://github.com/KimMeen/Time-LLM. The code for Moment is available at https://github.com/moment-timeseries-foundation-model. The code for Mantis is available at https://github.com/vfeofanov/mantis. Other models can be found on https://huggingface.co/models. 

\bibliographystyle{unsrt}
\bibliography{references}  






\end{document}